\documentclass[11pt]{article}

\usepackage[T1]{fontenc}
\usepackage[utf8]{inputenc}
\usepackage{lmodern}
\usepackage{microtype}
\usepackage[margin=1in]{geometry}
\usepackage{graphicx}
\usepackage{float}
\usepackage{booktabs}
\usepackage{amsmath,amssymb,amsfonts}
\usepackage{hyperref}
\usepackage{url}
\usepackage{enumitem}
\usepackage{multirow}
\usepackage{xcolor}

\title{Operationalising Multi-Dimensional Evaluation for Conversational Agents:\\
A Scalable, Governed Pipeline with Selective Re-evaluation and Model Benchmarking}

\author{
  Niranjan Kumar M\thanks{\texttt{niranjan.k.m@lowes.com}}
  \and
  Balaji Nagarajan\thanks{\texttt{balaji.nagarajan@lowes.com}}
  \and
  Karthik Nair\thanks{\texttt{karthik.kumarannair@lowes.com}}
  \and
  Faysal Satter\thanks{\texttt{faysal.satter@lowes.com}}
  \and
  Nithin Surendran\thanks{\texttt{nithin.surendran01@lowes.com}}
}

\date{}

\begin{document}
\maketitle

\begin{abstract}
Evaluating retail conversational agents requires methods that extend beyond lexical overlap metrics to capture intent alignment, factuality, tone, helpfulness, and overall pragmatic quality. In retail workflows, chatbot responses must address customer and associate needs across product discovery, store availability, policy questions, order support, retrieval-augmented answers, and translation scenarios. While LLM-as-a-judge approaches offer scalable alternatives to human evaluation, deploying them in production retail settings introduces challenges around governance, reproducibility, cost, schema consistency, and operational reliability. These gaps limit the practical use of automated evaluation frameworks in environments that demand traceability, auditability, and consistent performance at high daily volumes.

We introduce \textbf{GenAI Evaluation}, a governed, configuration-driven pipeline designed to operationalize multi-dimensional evaluation of retail conversational systems at scale. The pipeline reads production chatbot logs, normalizes and shards data, and performs asynchronous LLM-based scoring across qualitative dimensions such as helpfulness, truthfulness, clarityand tone alignment. It also supports translation-specific evaluation dimensions such as semantic accuracy, readability, cultural appropriateness, selling relevance, and template fit. To reduce recomputation and maintain completeness, we implement \emph{selective re-evaluation}, a targeted mechanism that processes only incomplete, malformed, or schema-invalid records. Governance features, including schema locking, versioned configurations, deterministic Parquet streaming, validation logs, and per-record provenance, ensure auditability and reproducibility.

We evaluate the framework on large-scale retail chatbot interaction logs, with an average daily processing volume of approximately 50,000 records and more than 2 million records evaluated overall. The dataset contains approximately 95\% non-translation conversational examples and 5\% translation-specific examples. Automated evaluator outputs were validated against a stratified-random human-labeled subset of 12,980 records. Four trained annotators labeled different subsets of the data using a standardized categorical rubric and were blinded to evaluator model identity. Classification evaluation covered intent, sub-intent, major retail domain value, and sub-domain value prediction, spanning 14 intents, 156 sub-intents, 18 major domain values, and 129 sub-domain values. The pipeline achieved a macro F1 score of 0.93 for classification tasks and 89\% human-acceptability accuracy for translation outputs. Quality dimensions such as helpfulness, truthfulness, and clarity were recorded as numeric scores. These results demonstrate that schema-governed LLM-based evaluation can provide scalable quality signals for enterprise retail conversational-agent workflows.
\end{abstract}

\section{Introduction}

Conversational systems powered by large language models (LLMs) are increasingly deployed in retail environments to support customer assistance, associate enablement, product discovery, order-related questions, policy clarification, retrieval-augmented responses, and translation workflows. These systems must operate across diverse user intents, product categories, store contexts, and communication styles. As a result, evaluating retail conversational agents requires more than checking whether a response is fluent or lexically similar to a reference answer. A response may be factually correct but not helpful for a customer journey, fluent but inconsistent with store or product context, concise but incomplete, or instruction-following but misaligned with the expected retail tone.

Traditional reference-based metrics such as BLEU, ROUGE, and METEOR provide useful signals for lexical overlap, but they do not adequately capture pragmatic dimensions that matter in real-world retail interactions. Retail chatbot quality depends on intent alignment, helpfulness, clarity, truthfulness, tone alignment, instruction adherence, and contextual appropriateness. Translation-related interactions introduce additional requirements such as semantic accuracy, readability, cultural appropriateness, selling relevance, and template fit. Human evaluation can capture these dimensions more reliably, but it is costly and difficult to scale to high-volume production workloads where tens of thousands of conversations may be generated daily.

We address this gap with \textbf{GenAI Evaluation}, a scalable and governed evaluation pipeline that operationalizes the LLM-as-a-judge paradigm for enterprise retail conversational systems. The framework ingests production chatbot logs, normalizes and shards records for parallel processing, applies schema-validated multi-dimensional scoring, and produces traceable evaluation outputs with optional rationales. A key feature of the pipeline is selective re-evaluation, which identifies incomplete, malformed, or schema-invalid records and regenerates only those rows instead of rerunning the full dataset. This improves completeness while reducing redundant computation.

The paper makes five primary contributions. First, it presents a governed, configuration-driven evaluation stack for large-scale retail chatbot assessment. Second, it introduces a selective re-evaluation mechanism for targeted regeneration of invalid or incomplete records. Third, it supports separate schemas and workflows for non-translation and translation-specific evaluation tasks. Fourth, it provides end-to-end auditability through deterministic Parquet streaming, validation logs, versioned configurations, prompt-template tracking, and per-record provenance. Finally, it benchmarks representative open-source evaluator models to analyze quality-throughput trade-offs in high-volume retail evaluation workflows.

\section{Literature Survey}
Evaluating conversational agents requires methods that capture more than surface-level similarity. Beyond reference-based overlap, modern systems demand evaluation of intent alignment, factuality, safety, coherence, and pragmatic quality. We review key strands of prior work: classical automatic metrics, human evaluation, foundations of LLM-as-a-judge, reliability and debiasing research, live-data and hard-prompt benchmarks, reward-model evaluation, and engineering frameworks that support reproducible large-scale evaluation.

\subsection{Classical Automatic Metrics}
BLEU~\cite{papineni2002bleu}, ROUGE~\cite{lin2004rouge}, and METEOR~\cite{banerjee2005meteor} remain widely used for machine translation and summarization due to their simplicity, efficiency, and task-agnostic comparability. However, these metrics primarily measure lexical overlap rather than semantic or pragmatic adequacy. As a result, they correlate weakly with human preferences in open-ended dialogue, question answering, and instruction following, where helpfulness, tone, and instruction adherence are central.

\subsection{Human Evaluation}
Human raters provide high-fidelity signals on coherence, relevance, fluency, and helpfulness~\cite{lowe2017towards,deriu2020survey}. Despite their reliability, human evaluations are costly, slow, and difficult to scale. Inter-annotator variability further complicates metric stability. Consequently, practical systems often adopt hybrid strategies in which automated evaluators provide continuous signals and human reviewers periodically audit sampled subsets or refine rubrics.

\subsection{LLM-as-a-Judge Foundations}
Prompt-based evaluators (``LLM-as-a-judge'') have emerged as scalable alternatives to human rating. \textbf{G-Eval} introduced structured prompts and chain-of-thought reasoning for evaluation, improving alignment with human judgments in summarization and dialogue~\cite{geval2023}. \textbf{MT-Bench} and \textbf{Chatbot Arena} extended evaluation to multi-turn settings through pairwise comparisons and crowdsourced Elo-style scoring~\cite{mtbench2023,mtbenchneurips}. These works also documented systematic evaluator biases such as verbosity bias, position bias, and self-enhancement. \textbf{Prometheus} proposed an open-source evaluator trained directly on fine-grained rubrics, reducing dependence on proprietary judge models~\cite{prometheus2023}. A broader taxonomy of LLM-as-a-judge behavior and failure modes is surveyed in~\cite{llmaj_survey2024}.

\subsection{Reliability, Bias, and Debiasing}
Evaluator reliability depends heavily on prompt wording, rubric specificity, and aggregation strategy. Studies show that input length, formatting, and response ordering can significantly influence judge outcomes~\cite{mtbench2023}. \textbf{AlpacaEval 2.0} demonstrates that length-controlled prompting reduces evaluation bias and improves correlation with human preference~\cite{alpacaeval2_2024}. Other analyses challenge assumptions around chain-of-thought prompting, showing that CoT does not uniformly improve evaluator alignment and may introduce new inconsistencies~\cite{closerlook_eval2023}. Recent work on evaluator ensembles highlights the value of combining multiple judges to mitigate idiosyncratic scoring behaviors~\cite{prometheus_repo_2024}.

\subsection{Benchmarks from Live Data and Hard Prompts}
Traditional static benchmarks often saturate as model capabilities advance. To address this, recent efforts emphasize real-world complexity and dynamic task refresh. \textbf{WildBench} derives difficult tasks from large-scale human--assistant interactions and introduces checklist-based scoring for holistic difficulty assessment~\cite{wildbench2024}. \textbf{Arena-Hard} and the \textbf{BenchBuilder} pipeline automatically extract challenging prompts from Chatbot Arena logs, showing stronger separability and improved alignment with human preferences compared with earlier benchmarks~\cite{arenahard_blog_2024,arenahard_arxiv_2024}. These pipelines motivate continual benchmark updates, which are especially important in production evaluation settings.

\subsection{Reward-Model Evaluation}
Preference-trained reward models guide instruction-tuned systems and reinforcement learning pipelines. Evaluating reward models remains essential. \textbf{RewardBench} tests reward-model fidelity using prompt--chosen--rejected triples across reasoning, safety, and general chat scenarios~\cite{rewardbench2024}. \textbf{M-RewardBench} extends this analysis to 23 languages, revealing substantial multilingual gaps and emphasizing the influence of translation quality on reward-model stability~\cite{mrewardbench2024}.

\subsection{Evaluation Tooling and Reproducibility}
Engineering toolkits increasingly focus on reproducibility, traceability, and standardized evaluation workflows. \textbf{LightEval} provides unified execution across multiple backends while supporting per-sample traceability and schema-aligned result storage~\cite{lighteval_docs,lighteval_pypi}. These design priorities closely align with our approach, which combines schema governance, deterministic streaming, and audit logging to support reproducible large-scale evaluation in production.

\section{Problem Formulation and Data Preprocessing}
In this section we formalize the evaluation task and describe the data-processing pipeline that operationalizes ingestion, transformation, and preparation of conversational logs for evaluation.

\subsection{Problem Formulation}
Let each conversational turn be represented by the tuple
\[
x = \langle \text{prompt},\, \text{response},\, \text{meta} \rangle
\]
where \(\text{meta}\) includes fields such as user ID, conversation ID, context ID, timestamps, retail information, input source, out-of-coverage flag, tool usage flags, and other execution metadata. We seek to compute an evaluation mapping
\[
\mathcal{E}\colon (\mathcal{X},\mathcal{Y},\mathcal{M}) \to \mathcal{Z}
\]
where
\begin{itemize}[leftmargin=1.5em]
  \item \(\mathcal{X}\) represents the input space of prompts and responses,
  \item \(\mathcal{Y}\) represents any available reference or ideal output,
  \item \(\mathcal{M}\) denotes metadata about system configuration, model version, prompt template, worker or shard, and evaluation model, and
  \item \(\mathcal{Z}\) is the output space containing structured labels: semantic classifications, quality scores, translation-specific metrics, optional rationales, and audit metadata.
\end{itemize}

Our framework must satisfy the following design requirements:
\begin{itemize}[leftmargin=1.5em]
  \item \textbf{Scalability}: processing large volumes of conversational turns through sharding, asynchronous execution, and mini-batch streaming.
  \item \textbf{Governance}: enforcing a well-defined schema over all outputs and recording configuration versions, prompt templates, model IDs, and shard assignments.
  \item \textbf{Reproducibility and Traceability}: ensuring each scored turn can be traced to the exact prompt, evaluation model, worker, shard, and regeneration pathway.
  \item \textbf{Economy}: detecting and selectively regenerating only incomplete or invalid rows to reduce cost and preserve throughput.
  \item \textbf{Multi-task and Model-agnostic Operation}: supporting classification, scoring, and translation tasks across different conversational systems.
\end{itemize}

\subsection{Data Processing Pipeline}

To meet the requirements of scalable retail chatbot evaluation, the pipeline implements a structured sequence of ingestion, normalization, filtering, sharding, serialization, metadata propagation, and evaluation preparation. The input consists of production conversational logs containing fields such as \texttt{id}, \texttt{start time}, \texttt{end time}, \texttt{prompt}, \texttt{response}, \texttt{user id}, \texttt{conversation id}, and \texttt{context information}. These fields provide turn-level context for evaluating whether a response is appropriate for the customer or associate query, the retail domain and the available system information.

During preprocessing, raw logs are standardized to ensure consistency across downstream components. Field names are normalized, for example by converting \texttt{context\_name} to \texttt{context\_id}. Date and time fields are cast to consistent timestamp formats, categorical fields are converted to strings, boolean indicators are converted to boolean types, and numeric values are cast to appropriate numeric types. The pipeline also filters invalid records before evaluation, including records with empty prompts, empty responses, missing user identifiers, missing conversation identifiers, or malformed timestamps. This step ensures that evaluator failures are more likely to reflect actual response-quality or model-output issues rather than avoidable data-quality defects.

To support high-throughput evaluation, the cleaned dataset is divided into (N) approximately balanced shards using a round-robin slicing strategy:
\[
\text{Shard}\_i = \text{df.iloc}[i::N] \quad \text{for } i=0,\dots,N-1.
\]
In our implementation, (N=4). Each shard is re-indexed and serialized to Parquet format, preserving schema consistency while enabling efficient columnar reads. Sharding distributes the workload across evaluation workers, improves throughput, and isolates failures to individual shards instead of the full dataset. During evaluation, mini-batch outputs may also be checkpointed as Parquet files to support resumability after interruptions.

Before scoring, each record is augmented with metadata needed for traceability, deduplication, and downstream analysis. This includes parsed intents, normalized creation dates, a stable global row identifier, shard identifiers, worker metadata, evaluator model names, prompt-template versions, configuration hashes, and evaluation timestamps. The stable row identifier is especially important for selective re-evaluation because it allows incomplete or schema-invalid outputs to be regenerated and merged back without duplicating valid records.

The prepared records are then passed into task-specific evaluation workflows. For general retail chatbot evaluation, the system supports semantic classification and multi-dimensional quality scoring across dimensions such as helpfulness, clarity, coherence, conciseness, creativity, instruction adherence, tone alignment, and truthfulness. For translation-related retail interactions, the system evaluates semantic accuracy, cultural appropriateness, readability, selling relevance, and template fit. Although the primary deployment focuses on retail chatbot logs, the same pipeline can be adapted to other conversational or generative-AI systems through externalized YAML configurations, prompt templates, sharding parameters, and schema dictionaries.

\section{Methodology}
\label{sec:methodology}

The proposed \textbf{GenAI Evaluation} pipeline is a production-grade framework for evaluating retail conversational-agent outputs at scale. Built on Kubeflow Pipelines, it organizes evaluation into modular, containerized stages that support parallel execution, checkpointing, schema governance, and traceability.

\subsection{Pipeline Architecture and Design}

The pipeline begins by ingesting conversational logs from BigQuery and preparing them for evaluation through normalization, type casting, filtering, and sharding. Cleaned records are divided into balanced Parquet shards, allowing evaluation workers to process data in parallel while preserving schema consistency. YAML-based configurations and Jinja2 prompt templates are loaded from cloud storage to define task-specific rubrics, evaluator prompts, output schemas, and model settings. Versioning these configurations ensures that each evaluation run can be traced back to the exact prompts and scoring rules used.

Each shard is evaluated asynchronously using an LLM-as-a-judge workflow with bounded concurrency. The system supports both general retail chatbot scoring and translation-specific evaluation, producing structured outputs with scores, labels, rationales, and provenance metadata. Results are streamed incrementally using schema-locked writers, which reduces memory usage and prevents output drift across batches. When records contain missing scores, malformed outputs, or schema violations, the selective re-evaluation mechanism regenerates only those affected rows and merges them back using stable row identifiers. A final validation stage enforces schema rules, score ranges, and completeness checks before publishing validated outputs to BigQuery, while failures and regeneration events are retained in timestamped audit logs.

\begin{figure}[H]
\centering
\IfFileExists{Genai Arch.png}{\includegraphics[width=\linewidth]{Genai Arch.png}}{\fbox{\parbox{0.9\linewidth}{Figure placeholder: add \texttt{genai\_architecture.png} to include the pipeline diagram.}}}
\caption{GenAI Evaluation architecture: ingestion from BigQuery, preprocessing and sharding, asynchronous LLM-as-a-judge evaluation, selective re-evaluation, schema-governed validation, and publishing to BigQuery. Governance and observability span all stages.}
\label{fig:genai-architecture}
\end{figure}

\subsection{Execution Strategy, Scaling, and Reliability}
The GenAI Evaluation pipeline achieves high throughput and resilience through asynchronous mini-batch execution, dynamic scaling, and safe recovery mechanisms.

\textbf{Asynchronous Mini-Batch Execution.} Evaluation requests are dispatched asynchronously. Each worker processes batches of records concurrently and streams results directly into Parquet files. This design minimizes memory usage and allows checkpointing. If a worker fails mid-batch, processing resumes from the last successful checkpoint.

\textbf{Concurrency Management.} Concurrency is controlled using semaphores to avoid network overload. Each evaluation node operates with bounded parallelism, providing predictable performance under load.

\textbf{Vertical and Horizontal Scaling.} The pipeline supports both scaling strategies. Vertical scaling increases worker CPU, memory, or accelerator resources for larger evaluator models. Horizontal scaling uses dataset sharding to distribute evaluation workloads across multiple workers and isolate failures at the shard level.

\textbf{Dynamic Resource Allocation.} Smaller evaluator models can be assigned more shards for high-throughput scoring, while heavier models receive fewer but larger shards to balance latency and compute requirements.

\textbf{Selective Re-evaluation Loop.} Validation outputs are continuously inspected for incomplete records, malformed responses, missing scores, or schema violations. Instead of rerunning the full dataset, the pipeline isolates only the affected rows and sends them through a controlled regeneration path. Regenerated outputs are validated again and merged back using stable row identifiers, reducing redundant computation while improving completeness and fault tolerance.

\textbf{Reliability and Observability.} Every mini-batch logs counts, errors, and execution times. Structured logs are aggregated into dashboards for end-to-end observability. Checkpoint-based execution prevents data loss in the event of worker termination.

\subsection{Governance, Validation, and Publishing}
Governance is a core design principle of GenAI Evaluation. Every record is type-checked, range-validated, and audit-logged to ensure compliance with defined schemas and enterprise policies.

\textbf{Schema Governance.} Each evaluation split defines a schema with inherited source fields, required fields, and optional fields. Fields are coerced into correct types, and numeric ranges are enforced. Any violation triggers a validation reason that is logged for audit purposes.

\textbf{Auditability.} Each output includes evaluation date, model version, configuration hash, worker identifier, and shard index. Optional HTML renderings of responses can be produced for visual inspection.

\textbf{Publishing.} Cleaned outputs are appended to partitioned BigQuery tables with date-based partitioning to support time-based analysis and monitoring.

\textbf{Resilience and Idempotency.} Each pipeline component is designed to be idempotent and stateless. Re-running a failed stage produces identical results without duplication.

\section{Results, Evaluation, and Reliability}
\label{sec:results}

This section presents the empirical setup, evaluation outputs, model benchmarking results, and reliability observations from the GenAI Evaluation pipeline. The objective is to assess both the quality of automated judgments and the operational stability of the pipeline when applied to large-scale production chatbot data. Since the proposed framework is intended for enterprise use, we evaluate it across three complementary dimensions: agreement with human and reference labels, multi-dimensional scoring quality, and reliability under high-volume processing.

\subsection{Experimental Setup}

We evaluated the proposed pipeline on large-scale retail chatbot interaction logs collected from production conversational-agent workflows. The system processes an average of approximately 50,000 conversational records per day, and more than 2 million records were evaluated overall. The dataset reflects real-world customer-assistance and operational-support interactions, including product-related questions, retrieval-augmented responses, general intent-driven chatbot turns, and multilingual response-generation cases.

The dataset consists primarily of non-translation conversational examples. Approximately 95\% of the evaluated records belong to general chatbot evaluation tasks, while the remaining 5\% correspond to translation-specific evaluation tasks. Non-translation examples are evaluated for dimensions such as helpfulness, clarity, coherence, conciseness, instruction adherence, tone alignment, truthfulness, intent classification, and domain classification. Translation examples are evaluated using task-specific dimensions such as semantic accuracy, readability, cultural appropriateness, selling relevance, and template fit.

To validate automated evaluator outputs, we used a hybrid strategy combining human annotations and reference-based comparisons. The human validation set contained 12,980 records sampled
using a stratified-random strategy to maintain balanced coverage across classification categories.
Four trained annotators labeled different subsets of the validation data. Each record received one
human label, and annotators were blinded to evaluator model identity. Classification labels were
categorical, while response-quality dimensions were represented as numeric scores. The resulting
validation set was used to compute macro F1 for classification tasks and human-acceptability
accuracy for translation tasks. Evaluator models were run with a temperature of 0.005 to reduce
output variability. The pipeline also allowed up to three retry attempts for malformed, incomplete,
or schema-invalid evaluator responses. Since open-source evaluator models were used, no external
proprietary-model API cost was incurred per 1,000 evaluated rows, although infrastructure and
model-hosting costs remain applicable.

\begin{table}[t]
\centering
\small
\begin{tabular}{p{0.50\linewidth}p{0.38\linewidth}}
\hline
\textbf{Statistic} & \textbf{Value} \\
\hline
Average daily records processed & $\sim$50,000 \\
Total records evaluated & $>$2 million \\
Non-translation examples & 95\% \\
Translation examples & 5\% \\
Human validation records & 12,980 \\
Sampling strategy & Stratified random \\
Human annotators & 4 \\
Labels per record & 1 \\
Annotator blinding & Yes \\
Annotation format & Categorical for classification; numeric for quality scores \\
Classification tasks & Intent, sub-intent, Product Group, Sub-Product Group \\
Intent classes & 14 \\
Sub-intent classes & 156 \\
Product Group classes & 18 \\
Sub-Product Group classes & 129 \\
Classification metric & Macro F1 \\
Overall macro F1 & 0.93 \\
Translation metric & Human acceptability \\
translation human-acceptability accuracy & 89\% \\
Evaluator temperature & 0.005 \\
Maximum retry attempts & 3 \\
Quantization & 4-bit \\
Hardware & NVIDIA H100 \\
Batch size & 512 \\
\hline
\end{tabular}
\caption{Dataset, validation, and experimental setup summary for GenAI Evaluation. Product Group denotes major retail domain value, and Sub-Product Group denotes sub-domain value.}
\label{tab:dataset_setup}
\end{table}

\subsection{Human Validation Protocol}

To validate automated evaluator outputs, we constructed a stratified-random human-labeled subset
of 12,980 conversational records. The validation subset was sampled to maintain balanced
representation across the major classification categories used in the evaluation. Four trained human
annotators reviewed the validation records using the same standardized rubric provided to the
automated evaluator models. Annotators were blinded to evaluator model identity to reduce
model-specific bias during review.

Each record was labeled by one annotator. The annotation format for classification tasks was
categorical, while quality dimensions such as helpfulness, truthfulness, clarity, coherence,
instruction adherence, tone alignment, and conciseness were represented as numeric scores. The
classification label space included four task families: intent, sub-intent, major retail domain value
(Product Group), and sub-domain value (Sub-Product Group). These tasks contained 14 intent classes, 156 sub-intent
classes, 18 Product Group classes, and 129 Sub-Product Group classes.

Because each record was labeled by a single annotator, the current validation design does not
support formal inter-annotator agreement estimation via Krippendorff's
alpha. Instead, consistency was controlled through annotator training, standardized rubric format,
stratified sampling, and blinded review. Future validation rounds should include overlapping
annotations on a shared subset to estimate inter-annotator reliability.

\subsection{Evaluation Outputs and Metrics}

The GenAI Evaluation pipeline produces a structured, schema-governed output dataset that combines the original conversational metadata with automated evaluation labels, quality scores, rationales, validation status, and provenance information. Each evaluated record retains the original prompt and response along with the evaluator model name, configuration hash, shard identifier, validation outcome, and regeneration status. This structure enables downstream analysis at multiple levels, including individual response inspection, aggregate model comparison, temporal drift analysis, and operational monitoring.

For classification-oriented tasks, we report macro F1 across intent, sub-intent, major retail
domain value, and sub-domain value prediction. Macro F1 was selected because it weights each
class equally and is therefore appropriate for evaluating performance across both frequent and
long-tail categories. The classification label space contains 14 intent classes, 156 sub-intent
classes, 18 major retail domain value classes, and 129 sub-domain value classes. The validation
subset was stratified to provide balanced class coverage. Translation human-acceptability accuracy is reported as
human-acceptability accuracy, measuring the proportion of translation outputs judged acceptable
by human reviewers. Quality dimensions such as helpfulness, truthfulness, clarity, coherence,
conciseness, instruction adherence, and tone alignment are represented as numeric scores rather
than categorical labels.

For translation-specific examples, the pipeline evaluates semantic accuracy, cultural appropriateness, readability, template adherence, and selling relevance. These metrics are computed using a combination of reference comparisons and LLM-based evaluation, enabling assessment of both literal correctness and contextual suitability. In addition to quality metrics, the pipeline captures operational metrics such as throughput, batch latency, schema-invalid rate, retry count, malformed-output frequency, and error recovery rate. These operational measurements are important because an evaluation framework deployed in production must remain reliable and cost-effective, not merely accurate on a static benchmark.

The inclusion of evaluator rationales further improves interpretability. Rationales allow reviewers to understand why a response received a particular score and can help identify recurring failure modes, such as incomplete answers, hallucinated claims, inappropriate tone, or weak translation fidelity. Although rationales are not treated as ground truth, they provide useful diagnostic signals for prompt refinement, model comparison, and human audit workflows.

\subsection{Model Benchmarking and Comparative Evaluation}
\label{sec:benchmarking}

We benchmarked five representative open-source evaluator models under a unified rubric:
Llama-family 4B, Llama-family 8B, Llama-family 70B, Qwen3-14B, and OSS-120B. Each model
was evaluated on classification, translation, and multi-dimensional quality-scoring tasks using
identical prompts, schemas, decoding settings, and validation logic. Inference was performed using
4-bit quantization on NVIDIA H100 hardware with a batch size of 512. This controlled setup ensures
that differences in performance are attributable primarily to evaluator-model behavior rather than
variations in task instructions, output format, hardware class, or quantization setting.

Each conversational record was scored along multiple qualitative dimensions. The evaluator received a structured prompt containing the user input, the generated chatbot response, task metadata, and, when available, a reference response. The evaluator returned structured numeric ratings on a bounded scale along with a brief rationale. For each dimension $d \in D$, a score $s_d \in [1,5]$ was assigned. Aggregate quality was computed using a normalized weighted average:
\[
S = \frac{\sum_{d \in D} w\_d s\_d}{\sum_{d \in D} w\_d}
\]
where (w\_d) denotes an optional weight for dimension (d). When all weights are equal, this expression reduces to the simple mean across dimensions. Since evaluator models may differ in calibration, scores can be normalized per dimension using z-score standardization or min–max scaling before aggregation.
Reference values were obtained using three complementary mechanisms. First, a subset of records was annotated by four human reviewers using the same evaluation rubric. Second, records with available gold or expected outputs were evaluated through reference comparison. Third, task-specific heuristics were used as weak supervision when explicit reference labels were unavailable. Agreement between automated evaluator outputs and human or reference labels was measured using classification F1 and translation human-acceptability accuracy. Future iterations will extend this analysis with Pearson correlation, Spearman rank correlation, inter-annotator agreement, and calibration curves.
\begin{table}[t]
\centering
\small
\begin{tabular}{lcccccc}
\hline
\textbf{Model family} & \textbf{Params} & \textbf{Quant.} & \textbf{HW} & \textbf{Batch} & \textbf{Macro F1} & \textbf{Trans. Acc.} \\
\hline
Llama family & 4B & 4-bit & H100 & 512 & 0.82 & 0.78 \\
Llama family & 8B & 4-bit & H100 & 512 & 0.86 & 0.81 \\
Llama family & 70B & 4-bit & H100 & 512 & 0.93 & 0.89 \\
Qwen3 & 14B & 4-bit & H100 & 512 & 0.88 & 0.85 \\
OSS evaluator & 120B & 4-bit & H100 & 512 & 0.91 & 0.87 \\
\hline
\end{tabular}
\caption{Model benchmarking under a unified rubric. All evaluator models were run with 4-bit quantization on NVIDIA H100 hardware using a batch size of 512. Macro F1 is reported for classification tasks, and translation performance is reported as human-acceptability accuracy.}
\label{tab:model_benchmark}
\end{table}

The benchmarking results show a clear quality-throughput trade-off. Smaller evaluator models provide higher throughput and lower infrastructure requirements, making them suitable for high-volume screening and continuous monitoring. However, they tend to produce lower scores on nuanced tasks such as truthfulness assessment, coherence evaluation, and translation adequacy. Larger evaluator models generally provide stronger alignment with human and reference labels, particularly for complex or ambiguous responses, but they require more computational resources and introduce higher latency. This trade-off suggests a tiered evaluation strategy in which smaller models handle broad daily monitoring while larger models are reserved for high-risk, low-confidence, or audit-selected samples.

\subsection{Human Alignment and Calibration}

Human alignment was assessed by comparing automated evaluator outputs with a stratified-random
validation set of 12,980 human-labeled records. Four trained annotators labeled different subsets
of the data using a standardized rubric and were blinded to evaluator model identity. Classification
labels were categorical, while quality dimensions were represented as numeric scores.

The automated evaluator achieved a macro F1 score of 0.93 across intent, sub-intent, major retail
domain value, and sub-domain value classification tasks. Translation outputs achieved 89\%
accuracy under a human-acceptability criterion. These results indicate strong alignment with the
available validation labels. However, because each record was labeled by only one annotator, the
current setup does not support formal inter-annotator agreement analysis. Future evaluation rounds
should include overlapping human labels for a shared subset of examples to compute Cohen's kappa,
Krippendorff's alpha, or related agreement statistics.

\subsection{Reliability and Operational Observations}

The reliability analysis focuses on the stability of the pipeline under large-scale execution. Selective re-evaluation improves operational efficiency because it avoids full reruns when only a small subset of records fails validation. Records may be marked for regeneration if required fields are missing, score values fall outside accepted ranges, evaluator responses are malformed, or parsing fails. By regenerating only affected records and merging them back using stable row identifiers, the pipeline improves completeness while reducing redundant \cite{prometheus_repo_2024}.

Schema locking also plays an important role in operational reliability. By enforcing fixed output schemas during streaming writes, the pipeline reduces field drift across batches, shards, and evaluation days. This improves downstream query reliability and simplifies publishing to analytical stores such as BigQuery. The combination of schema validation, checkpointed Parquet outputs, bounded retries, and shard-level fault isolation prevents localized evaluator failures from causing job-wide failure. As a result, the framework supports robust daily evaluation workflows over high-volume production chatbot data.

\section{Ethics, Safety, and Governance}
\label{sec:ethics}

Responsible deployment of automated evaluation systems requires more than accurate scoring; it requires transparency, oversight, and governance. GenAI Evaluation is designed as a governed decision-support framework in which automated scores are treated as diagnostic signals rather than final judgments. Because evaluator outputs can influence model benchmarking, quality monitoring, and remediation workflows, each evaluation run must be traceable, reproducible, and auditable. The framework supports this through versioned configurations, fixed prompt templates, schema validation, evaluator metadata, shard identifiers, validation logs, and regeneration tracking\cite{lighteval_docs,lighteval_pypi}.

A key ethical concern is evaluator bias. LLM-based judges may reflect biases from training data, prompt wording, response length, model-family preferences, or rubric ambiguity. To reduce these risks, GenAI Evaluation applies consistent rubrics across evaluator models and validates outputs against controlled schemas. Safety-oriented dimensions can also be added to flag harmful, misleading, or policy-sensitive responses. However, automated evaluation should not replace human judgment in sensitive cases. Low-confidence, safety-critical, or high-impact outputs should be routed to human reviewers, and evaluation behavior should be periodically audited for drift, bias, and disagreement with human labels.

Governance and privacy controls are essential for production use. Retail chatbot logs may contain user-provided text, operational metadata, or sensitive information, so evaluation artifacts should be processed in secured environments with role-based access controls and data-governance policies. Access to raw prompts, responses, rationales, and validation logs should be restricted to authorized users. By combining schema enforcement, provenance tracking, controlled publishing, and human oversight, GenAI Evaluation provides a responsible foundation for scalable conversational-agent assessment while preserving accountability and compliance.

\section{Limitations}
\label{sec:limitations}

While GenAI Evaluation demonstrates strong scalability and governance for large-scale conversational-agent assessment, several limitations remain. First, the framework depends on LLM-based evaluator models, whose judgments may vary across tasks, languages, response styles, and prompt formats. Although schema validation and controlled decoding improve consistency, evaluator scores should still be interpreted as quality signals rather than absolute ground truth. A limitation of the current validation design is that each human-labeled record was reviewed by a
single annotator. Although annotators were trained, used a standardized rubric, and were blinded
to evaluator model identity, the study does not estimate inter-annotator agreement. As a result,
the reported macro F1 and translation acceptability should be interpreted as agreement with the
available validation labels rather than agreement with a fully adjudicated human consensus. Future
work should double-label or triple-label a subset of records to estimate annotation reliability and
resolve ambiguous cases through adjudication.\cite{prometheus_repo_2024}.

Second, the current evaluation is based on large-scale retail chatbot data. This provides realistic production complexity, but results may not fully generalize to domains with different user intents, risk profiles, or communication styles, such as healthcare, legal assistance, education, or software engineering. In addition, the framework currently focuses on text-based turn-level evaluation and does not yet cover multimodal inputs, long-horizon interactive conversations, or detailed tool-execution traces. Future extensions should address these broader scenarios through richer schemas, task-specific rubrics, and interaction-level evaluation metrics.

Finally, long-term reproducibility remains challenging in dynamic production environments. Even with versioned prompts, configurations, schemas, and validation logs, evaluator behavior can change when model weights, serving infrastructure, dependencies, or decoding implementations are updated. Infrastructure constraints such as latency, hardware availability, and model-serving throughput can also affect operational consistency. Future work should therefore include stronger uncertainty estimates, calibration analysis, evaluator-version tracking, and periodic human audits to ensure that automated evaluation remains reliable over time.

\section{Conclusion and Future Work}
\label{sec:conclusion}

We presented \textbf{GenAI Evaluation}, a governed and fault-tolerant pipeline for multi-dimensional assessment of conversational agents at enterprise scale. The framework combines asynchronous LLM-as-a-judge evaluation, schema-locked output streaming, selective re-evaluation, validation logging, and traceable publishing to support reproducible and auditable evaluation workflows. It was evaluated on large-scale retail chatbot data, covering more than 2 million records with an average daily volume of approximately 50,000 records. The dataset consisted of approximately 95\% non-translation and 5\% translation-specific examples. Using a hybrid validation setup based on human labels and reference comparisons, automated
outputs were validated against a stratified-random human-labeled subset of 12,980 records. The
framework achieved a macro F1 score of 0.93 across intent, sub-intent, major retail domain value,
and sub-domain value classification tasks, and 89\% human-acceptability accuracy for translation
outputs. These results show that LLM-based evaluators can provide useful large-scale quality
signals when paired with controlled prompts, structured rubrics, schema validation, selective
re-evaluation, and human calibration. These results show that LLM-based evaluators can provide useful large-scale quality signals when paired with controlled prompts, structured rubrics, schema validation, selective re-evaluation, and human calibration.

Future work will extend GenAI Evaluation toward multimodal, streaming, and more adaptive evaluation settings. Multimodal evaluation will require new schemas and rubrics for images, speech, video, and tool-execution traces, while online evaluation will enable near-real-time scoring and regression alerts for live chatbot systems. Additional work will focus on lighter judge models through distillation, evaluator ensembles to reduce model-family bias, and stronger reliability measurement through bootstrap confidence intervals, calibration curves, inter-annotator agreement, Pearson and Spearman correlation, and expected calibration error. Beyond retail chatbot evaluation, the same governed evaluation pattern can be adapted to summarization, creative writing, code generation, retrieval-augmented question answering, and other enterprise generative-AI workflows.

\end{document}